# 基于改进 YOLOX 的道路目标检测算法研究

杨涛，吴友宇，汤杨心泰

武汉理工大学

**摘要**：道路目标检测是自动驾驶技术中的一个重要分支，检测精度越高的模型越有利于车辆的安全驾驶。在道路目标检测中，小目标和遮挡目标的漏检是一个重要的难题，降低目标的漏检率对于安全驾驶具有重要意义。

在本文的工作中，基于 YOLOX 目标检测算法进行改进，提出了 DecIoU 边界框回归损失函数来提高预测框和真实框的形状一致性，并引入 Push Loss 来进一步优化边界框回归损失函数，以检测出更多的遮挡目标。此外，还使用了动态锚框机制来提升置信度标签的准确性，改善了无锚框目标检测模型的标签不准确的问题。

在 KITTI 数据集上的大量实验证明了所提出的方法的有效性，改进的 YOLOX-s 在 KITTI 数据集上的 mAP 和 mAR 分别达到 88.9%和 91.0%，相比基线版本提升 2.77%和 4.24%；改进的 YOLOX-m 的 mAP 和 mAR 分别达到 89.1%和 91.4%，提升了 2.30%和 4.10%。

**关键词**：Object Detection; YOLOX; DecIoU; Push Loss; Dynamic Anchor

## 1 介绍

近年来，自动驾驶汽车不断走进我们的视野中，面向自动驾驶的目标检测算法也成为了国内外的研究热点之一。安全可靠的自动驾驶汽车依赖于对周围环境的准确感知，以便及时做出正确的决策。目标检测是自动驾驶系统的关键任务之一，其主要的功能是检测前方道路上出现的目标的空间位置和目标类别[1]。

传统目标检测算法依赖于手工设计好的特征来对目标进行特征提取，以实现分类和检测的目的，常见的目标特征包括 Scale Invariant Feature Transform (SIFT)[2]、Speeded up robust features (SURF)[3]、histogram of oriented gradient (HOG)[4] et al. 该类方法设计出的特征泛化能力弱、鲁棒性较差。2012 年，Krizhevsky et al.提出了以 Convolutional Neural Networks (CNN)为基础的 AlexNet[5]算法框架，极大的提升了算法的速度和准确度。

相比于传统目标检测算法，以 CNN 为核心的目标检测算法具有准确率高、检测速度快等优点，发展潜力巨大。根据神经网络的结构不同，可以将目标检测算法分为两阶段目标检测算法和一阶段目标检测算法。2014 年，Girshick et al.提出了 R-CNN[6]目标检测算法，在目标检测数据集 VOC2012 上取得了 30%以上的精度提升。2015 年，基于 R-CNN 改进的 Fast R-CNN[7]、Faster R-CNN[8]等在检测速度和精度上获得进一步提升，逐渐成为了目标检测的首选方法。R-CNN 系列目标检测算法是典型的两阶段目标检测算法，第一阶段通过算法生成候选区域，第二阶段利用 CNN 网络对候选区域进行特征提取并根据提取的特征进行分类工作，得到最终的检测结果。

R-CNN 系列目标检测算法以较慢的速度换取

了较高的检测精度，在自动驾驶、智慧交通等实时检测场景中无法满足需求。2016 年，Redmon et al.提出了 You Only Look Once(YOLO)[9]有效的改善了这一问题。YOLO 网络只需要"看"一次输入图片，即可输出最终的检测结果，是典型的一阶段目标检测算法，具有网络结构简单、检测速度快等优点。随后，YOLO v2~v5[10-13]相继提出，较好的平衡了精度和速度的，YOLO 算法的优异性能表现将一阶段目标检测算法推向了主流。

2021 年，旷视科技的 Ge et al.进一步研究了 YOLO 系列目标检测算法，并融合了解耦头、AnchorFree[14-16]、SimOTA[17]和多正例等技术，提出了 YOLOX[18]目标检测算法，在满足实时性的前提下，进一步提升了近两个百分点的精度。

在本文的研究中，基于 YOLOX 目标检测算法进行损失函数的优化，以改善遮挡目标和小目标等困难目标检测精度较低的问题。简而言之，本文的主要贡献如下：

a) 提出 DecIoU，通过对面积进行解耦来优化预测框的形状，提高预测框和真实框的形状一致性，并与 IoU、GIoU、DIoU 等其他损失函数对比，证明了 DecIoU 的有效性；

b) 采用 Push Loss 并应用于边界框回归损失中，提高了 YOLOX 在 KITTI 数据集上的检测精度，检测出更多的遮挡目标；

c) 采用动态锚框来优化置信度标签分配，生成更准确的标签值以优化模型训练，最终得到检测性能更好的模型。

## 2 YOLOX

Redmon et al.提出的 YOLO 算法把目标检测问题转化为一个回归问题，通过一个统一的网络对目标实现分类和定位，只需要在输入端输入一张图像即可在输出端得到该图像中待检测目标的位置、类别和置信度。其主要的思想是将输入图片分割成 $S \times S$ 个网格,如果图像中目标中心落在了一个网格内，那么该网格就负责检测这个目标。每一个网格都会预测出 B 个边界框，每个边界框包含位置信息 $(x, y, w, h)$、置信度以及 N 个类别的概率，对于输出层来说，最终的输出维度为：$S \times S \times B \times (4+1+N)$ 的张量。

YOLOX 以 YOLO 系列目标检测算法为基础进行优化，选用 CSPDarkNet 作为主干网络，并采用了非线性表达能力更强的 SiLU 函数作为激活函数，增强了整个网络的学习能力。在检测头部分，YOLOX 提出了解耦头以解决目标检测中分类和回归冲突的问题。此外，YOLOX 实现了无锚框的目标检测，摆脱了 YOLO 系列检测器的锚框机制，将每个位置从预测 3 个边界框减少为预测 1 个边界框，并使它们直接预测偏移值，这种改进不仅减少了检测器的参数量和计算量，同时获得了更好的性能。在目标框匹配上，YOLOX 提出了先进的 SimOTA 匹配策略，这使得网络的训练时间有所减少，检测精度更加优异，展示了高级分配策略的优越性。YOLOX 主要包括 YOLOX-s、YOLOX-m、YOLOX-l、YOLOX-x 等版本，本文将以 YOLOX-s 为基线进行研究。

损失函数是神经网络的核心驱动力，在神经网络的训练过程中通过最小化神经网络的全局损失来优化整个网络的性能表现。YOLOX 的损失函数包含三个部分：分类损失、边界框回归损失和置信度损失。YOLOX 使用二元交叉熵损失函数来计算类别概率和目标置信度得分的损失，使用 IoU 或 GIoU 等损失函数来作为边界框回归的损失，总损失函数如下：

$$L_{YOLOX} = L_{cls} + L_{reg} + L_{obj} \quad (1)$$

# 3 本文方法

## 3.1 解耦 IoU 损失

目标检测任务可分为目标分类和目标定位两个任务。目标分类是要对检测到的目标进行分类以确定其属于哪一个类别。目标定位是要在图像中确定待检测目标的位置信息，输出其在图像中的坐标。目标定位依赖于边界框回归去定位目标，通过在模型训练过程中最小化边界框回归损失，以优化所预测边界框的位置，达到定位目标的目的。传统的边界框回归损失一般通过 L1 或 L2 距离范数来定义，忽视了坐标间的关联性。2016 年，Yu et al.在人脸检测任务中提出了 Intersection over Union(IoU)损失函数以建立坐标之间的关联性，提升边界框回归性能[19]。IoU 是比较两个形状之间相似性的最常用度量，是目标检测任务中的主要评价指标之一，将度量本身作为优化的目标是更佳的选择，IoU 损失已经在检测、跟踪和分隔等任务中广泛应用，成为边界框回归任务的最佳损失函数之一，IoU 和基于 IoU 的损失定义如下[20]：

$$IoU = \frac{|B \cap B^{gt}|}{|B \cup B^{gt}|} \quad (2)$$

$$L = 1 - IoU + R(B, B^{gt}) \quad (3)$$

式中 $B$ 表示预测框，$B^{gt}$ 表示真实框。然而当预测框和真实框不重合时 IoU 为 0，使用 IoU 损失将无法度量预测框和真实框的远近，无法进一步优化预测框。为了改善该问题，斯坦福学者 Rezatofighi et al.在 2019 年提出了 GIoU[21]，随后 Zheng et al.提出 DIoU[20]再一次优化了边界框回归损失函数，GIoU 和 DIoU 定义如下：

$$GIoU = IoU - \frac{|C/(B \cup B^{gt})|}{|C|} \quad (4)$$

$$DIoU = IoU - \frac{(b - b^{gt})^2}{c^2} \quad (5)$$

式中 $C$ 表示能包含 $B$ 和 $B^{gt}$ 的最小外接矩形框，$b$ 和 $b^{gt}$ 分别表示 $B$ 和 $B^{gt}$ 的中心点，$c$ 表示 $C$ 的对角线长度。

IoU 等损失函数主要从边界框面积之间的差距进行优化，在优化过程中无法保证预测框和真实框形状的相似性。受 L1 和 L2 损失函数的启发，我们在 IoU 损失基础上对边界框面积进行解耦，添加宽和高惩罚项，在最小化预测框和真实框面积差距的同时优化其形状相似性，这对于遮挡目标和小目标等困难目标检测有重要意义，更合理的检测框形状将减小该框在后处理过程中被过滤掉的概率，提升目标检测的召回率。本文将解耦 IoU 定义如下：

$$DecIoU = IoU - \frac{(C_w - I_w)^2}{C_w^2} - \frac{(C_w - I_h)^2}{C_h^2} \quad (6)$$

式中，$I_w$、$I_h$ 分别表示 $B$ 和 $B^{gt}$ 重叠部分的宽和高，$C_w$、$C_h$ 分别表示 $C$ 的宽和高。由此可得如算法 1 所示的 DecIoU 损失函数，以优化边界框回归。

---

**Algorithm 1**　DecIoU loss function

**input:**　$B = (x_1, y_1, x_2, y_2)$　as the predicted box,
　　　$B^{gt} = (x_1^{gt}, y_1^{gt}, x_2^{gt}, y_2^{gt})$　as the ground-truth.

**output:** $L_{DecIoU}$.

1. Calculating area of $B$： $A = (x_2 - x_1) \times (y_2 - y_1)$.
2. Calculating area of $B^{gt}$：
   $A^{gt} = (x_2^{gt} - x_1^{gt}) \times (y_2^{gt} - y_1^{gt})$.
3. Calculating intersection $I$ between $B$ and $B^{gt}$：
   $x_1^I = \max(x_1, x_1^g)$, $x_2^I = \min(x_2, x_2^g)$,
   $y_1^I = \max(y_1, y_1^g)$, $y_2^I = \min(y_2, y_2^g)$.
4. Finding the coordinate of smallest enclosing box $C$.
5. Calculating area of $C$： $A^c = (x_2^c - x_1^c) \times (y_2^c - y_1^c)$.
6. Calculating *iou* between $B$ and $B^{gt}$：
   ***if*** $x_2^I > x_1^I$ ***and*** $y_2^I > y_1^I$ ***then***：
   　　$I_w = (x_2^I - x_1^I)$, $I_h = (y_2^I - y_1^I)$.
   ***else***：

$I_w = 0$, $I_h = 0$.

7.  Calculating $iou$ and $deciou$:

$$C_w = (x_2^c - x_1^c), \quad C_h = (y_2^c - y_1^c),$$

$$iou = \frac{I_w \times I_h}{A + A^{gt} - I_w \times I_h},$$

$$deciou = iou - \frac{(C_w - I_w)^2}{C_w^2} - \frac{(C_h - I_h)^2}{C_h^2}.$$

8.  $L_{DecIoU} = 1 - deciou$.

### 3.2 Push-IoU 和 Push-DecIoU 损失

目标遮挡是目标检测任务的难题之一，Luo et al.提出 Pull loss 和 Push loss 应用于模型训练中，较好的改善了遮挡目标的漏检问题[22]。在目标检测算法训练的过程中，我们把预测框和真实框（Ground-truth）进行匹配，当一个预测框和 Ground-truth 匹配后，则该框为正样本；反之，未成功匹配的预测框为负样本。在 YOLOX 中，每个预测框最多匹配一个 Ground-truth，当道路上两个目标之间的发生遮挡时，相应的真实框之间出现部分重叠，这将使得两个目标最终的预测框之间出现重叠，在算法的后处理过程中有可能将重叠的预测框过滤掉，从而产生目标的漏检，每一个漏检的目标都关乎着车辆的行驶安全。为了进一步减小漏检情况的发生，我们对 IoU 损失进行了优化，改进后的 Push-IoU 损失函数包含 IoU 损失和 Push 损失两部分，如算法 2 所示。

---

**Algorithm 2**  Push-IoU loss and Push-DecIoU loss

**input:**  $B = (x_1, y_1, x_2, y_2)$ as the predicted box,

$B^{gt} = (x_1^{gt}, y_1^{gt}, x_2^{gt}, y_2^{gt})$ as the ground-truth,

$G$ as the set of all real boxes.

**output:**  $L_{Push-IoU}$, $L_{Push-DecIoU}$.

1.  Calculating $iou$ and $deciou$ between $B$ and $B^{gt}$:

$iou = IoU(B, B^{gt})$, $deciou = DecIoU(B, B^{gt})$.

2.  Finding the second ground-truth: find a real box $B^{gt`}$ which $IoU(B, B^{gt`})$ is the smallest, where $B^{gt`} \in G$ and $B^{gt`} \neq B^{gt}$.

3.  Calculating $iou$ between $B$ and $B^{gt`}$:

$iou` = IoU(B, B^{gt`})$.

4.  $L_{Push-IoU} = 1 - iou + \alpha \times iou`$,

$L_{Push-DecIoU} = 1 - deciou + \alpha \times iou`$.

---

在 Push 损失中，我们提出了"Second Ground-truth"，如图 3.1 所示，对于一个已经匹配了真实框的正样本预测框，将进一步在该预测框周围的所有真实框中寻找一个与之 IoU 最大的真实框作为"Second Ground-truth"（如图 3.1 中 gt`）。在训练的过程中，最大化该正样本预测框和 gt 框的 IoU，最小化该框和 gt`的 IoU，尽可能的将两个遮挡目标对应的预测框推开，减小重叠部分，降低在后处理过程中被过滤掉的可能性。此外，我们为 Push 损失设置了超参数 $\alpha$ 来调节 IoU 损失和 Push 损失的比例，以控制推开预测框的力度，避免预测框偏移过多而成为低质量预测框。

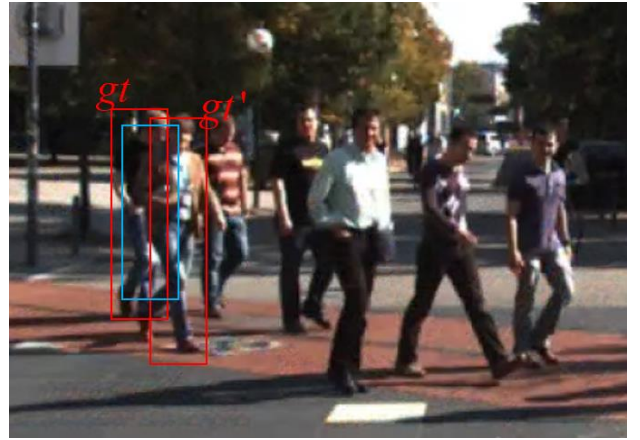

Figure 3.1 Second Ground-truth

### 3.3 动态锚框

置信度损失是目标检测损失函数中的重要损失之一，置信度是指边界框中包含目标的可能性

以及包含目标情况下边界框准确度的乘积，计算公式如下：

$$Confidence = P_{object} \times IoU_{pred}^{gt} \qquad (7)$$

预测框置信度的大小反应了该预测框中包含待检测目标的概率。在目标检测算法中，通常将输入图片分割成 $S \times S$ 个格子，如果一个格子中包含待检测目标，则 $P_{object}=1$，反之 $P_{object}=0$。对于基于锚框的目标检测算法，先验锚框通过训练集统计得出，可以较好的反应数据集中目标宽高的分布，在训练早期能够得到更加准确的预测框。YOLOX 是无锚框的目标检测算法，训练早期的预测框随机性较大，$IoU_{pred}^{gt}$ 较小，这将使得包含目标的单元格的置信度标签 $Confidence_{gt}$ 偏小，无法准确反应该单元格包含目标的概率。神经网络的训练是追求预测值和标签值的不断靠近，标签值的准确性对于目标检测模型的训练至关重要[23]。本文引入了动态锚框来辅助 $IoU_{pred}^{gt}$ 的计算，以生成更加准确的置信度标签值。如图 3.2 所示，$b_{pred}$ 和 $b_{gt}$ 分别为预测框和真实框的中心点，预测框的中心点已经较好的贴合真实框中心点，具有成为高质量预测框的潜力，然而由于宽和高的差距，最终 $IoU_{pred}^{gt}$ 和 $Confidence$ 较小，该预测框在后续迭代训练过程中可能会被逐渐忽略。本文以预测框的中心点 $b_{pred}$ 作为中心点，构建一个宽和高分别为 $w_{gt}$、$h_{gt}$ 的动态锚框（中心点随着预测框的动态变化而不断更新），以生成宽和高更加准确的辅助框用于计算置信度标签值，推动该预测框优化成为高质量预测框。

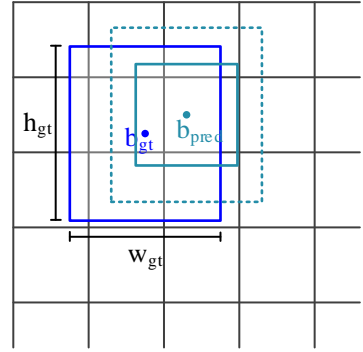

Figure 3.2 Dynamic Anchor

## 4 实验

### 4.1 实验环境

本文中改进的 YOLOX 算法基于 Pytorch1.11 和 CUDA11.5 进行训练和验证，实验平台配置如表 4.1，模型训练参数设置如表 4.2。

Table 4.1 实验平台配置

| Name | Parameter |
| --- | --- |
| System | Windows10 |
| CPU | Intel(R) Core(TM) i9-9900K |
| GPU | NVIDIA GeForce RTX 2080Ti |
| RAM | 32.0GB |

Table 4.2 模型训练参数

| Parameter | Value |
| --- | --- |
| learning rate | 0.01 |
| epochs | 500 |
| batch_size | 16 |
| optimizer_type | SGD |
| momentum | 0.937 |
| weight_decay | 0.0005 |

### 4.2 KITTI 数据集

KITTI 数据集是基于真实场景建立的一个庞大的自动驾驶数据集，包含市区、乡镇和高速公路等

场景的图像数据，可用于目标检测任务的模型训练和评估。KITTI 数据集包含 7481 张带有标签的图片，分为 Pedestrian、Truck、Car、Cyclist、DontCare、Misc、Van、Tram、Person_sitting 等 7 个类别。KITTI 数据集中不同类别样本数量极不均衡，可能导致模型无法收敛，本文进一步将 7 类目标划分为 Car、Cyclist、Pedestrian 三类，在训练过程中将 7481 张图片按照 7:1:2 的比例划分为训练集、验证集和测试集。

### 4.3 实验结果与分析

#### 4.3.1 与基线版本对比

本文在 YOLOX 中将边界框回归损失函数由 IoU 损失替换为 DecIoU 损失，我们还将与 GIoU、DIoU 在 YOLOX 中的表现进行比较，以评估我们提出的边界框回归损失函数。

在实验中，我们通过计算在特定 IoU 阈值下(本文取 0.5) 不同类别目标的 mean Average Precision(mAP)和 mean Average Recall(mAR)来评估模型的性能。同时，为了更全面的表达模型的检测能力，我们借鉴 COCO 数据集的评价指标，采用 $AP_S$、$AP_M$、$AP_L$、$AP_{0.5:0.95}$ 等指标进行综合分析，并在表 4.3 中报告了我们的实验结果。通过对比实验表明，GIoU、DIoU 作为回归损失可以稍微提高 YOLOX 在 KITTI 数据集上性能。而本文提出的 DecIoU 表现出更好的性能，实现了 88.4%的 mAP 和 89.6%的 mAR，相比于基线版本分别提升了 2.15% 和 2.63%。在 $AP_S$、$AP_M$ 和 $AP_{0.5:0.95}$ 等指标上 DecIoU 也有一定优势，这表明 DecIoU 通过对面积进行解耦，所预测的边界框更加准确，定位能力更强，尤其是对于中小目标有明显提升。图 4.1 展示了训练过程中，不同边界框回归损失的性能对比，在训练前期 DecIoU 与 GIoU 检测性能基本一致，相比 DIoU

收敛更慢，在训练 180 epoch 后，DecIoU 的检测性能逐渐超越 DIoU 和 GIoU 等，并最终在 400 epoch 后模型收敛，DecIoU 在四种边界框回归损失中表现最佳，证明了 DecIoU 的有效性。

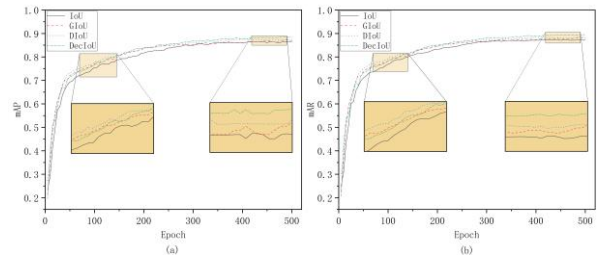

Figure 4.1 mAP and mAR of training

Table 4.3 边界框回归损失函数性能对比

| Loss | mAP | mAR | $AP_S$ | $AP_M$ | $AP_L$ | $AP_{0.5:0.95}$ |
|---|---|---|---|---|---|---|
| IoU | 86.5 | 87.3 | 42.2 | 58.7 | 69.3 | 58.3 |
| GIoU | 86.8 | 88.1 | 42.2 | 58.8 | 69.1 | 58.3 |
| improv% | 0.35 | 0.92 | 0.00 | 0.17 | -0.29 | 0.00 |
| DIoU | 87.3 | 88.3 | 42.7 | 58.5 | 69.3 | 58.4 |
| improv% | 0.92 | 1.15 | 1.18 | -0.34 | 0.00 | 0.17 |
| DecIoU | 88.4 | 89.6 | 42.9 | 59.8 | 69.5 | 59.1 |
| improv% | 2.15 | 2.63 | 1.66 | 1.87 | 0.29 | 1.37 |

Table 4.4 Results of PushLoss

| Loss | mAP | mAR |
|---|---|---|
| IoU | 86.5 | 87.3 |
| Push-IoU | 87.6 | 88.7 |
| Push-DecIoU | 88.3 | 90.5 |
| improv % | 2.08 | 3.67 |

进一步的，我们验证了 Push-IoU 损失和 Push-DecIoU 损失的检测性能，通过表 4.4 可以发现 Push-DecIoU 表现出最佳的检测性能，缓解了目标漏检的问题，在和 IoU 保持基本一致的 mAP 下，mAR 达到 90.5%，能够检测出更多的遮挡目标。如图 4.2 所示的相同图片上的检测结果，IoU Loss 检测出 10 个目标而 Push-IoU Loss 检测出 14 个目标，对于 IoU 漏检的遮挡目标，Push-IoU 成功的将其检测出来。

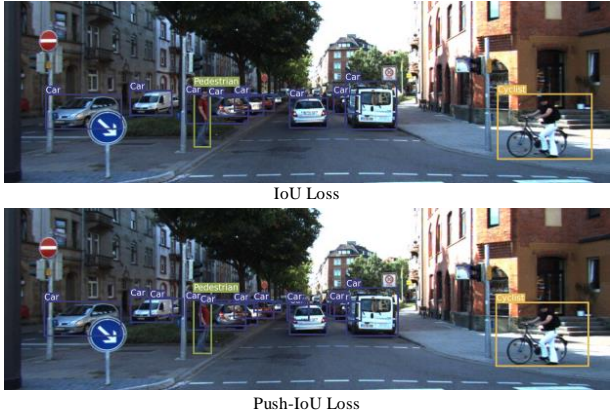

Figure 4.2 detection result of IoU Loss and Push-IoU Loss

### 4.3.2 消融实验

最终我们进行了多次消融实验来分析本文提出的改进方法，消融实验结果如表 3.5 所示。为了提升数据的可信度，我们基于 YOLOX-s 和 YOLOX-m 两个版本进行了实验，所有的实验数据均为 3 次实验的平均值。

正如表 4.5 的实验结果所示，本文提出的改进方法对 YOLOX-s 和 YOLOX-m 两个版本均有一定的改善，对 YOLOX 的提升是稳定的，mAP 分别提高 2.77%和 4.24%，mAR 分别提高 2.30%和 4.10%，通过消融实验证明了本文方法的有效性。

Table 4.5 消融实验结果

| Backbone | YOLOX-s | | YOLOX-m | |
|---|---|---|---|---|
| | mAP | mAR | mAP | mAR |
| —— | 86.5 | 87.3 | 87.1 | 87.8 |
| +DecIoU Loss | 88.4 | 89.6 | 88.7 | 90.0 |
| +Push-DecIoU Loss | 88.3 | 90.5 | 88.7 | 90.9 |
| +Dynamic Anchor | 88.9 | 91.0 | 89.1 | 91.4 |
| improv % | 2.77 | 4.24 | 2.30 | 4.10 |

## 5 结论

本文以 YOLOX 目标检测模型为基础，提出了 DecIoU 边界框回归损失函数和 Push-IoU 损失、Push-DecIoU 损失，对 YOLOX 的损失函数进行改进，并在 KITTI 数据集上进行实验验证。实验结果表明，改进的损失函数提升了模型的精确度同时降低了漏检率，mAP 和 mAR 分别提升了 2.08%和 3.67%。我们进一步对 YOLOX 的置信度标签计算方法进行了研究，通过提出动态 Anchor 辅助标签计算，优化了置信度标签的准确度，通过更准确的标签训练出检测能力更强的模型。本文改进的 YOLOX 在 KITTI 数据集上最终达到 89.1%的 mAP 和 91.4%的 mAR，对于道路目标的检测达到了较高的精度，有助于提升智能汽车的行车安全，具有一定的实用价值。

然而，本文也有一些不足之处。我们仅使用 KITTI 数据集进行了实验验证，下一步有待尝试更多更大的自动驾驶数据集进行实验研究，以不断改进我们的方法，提升对于道路目标的检测精度。此外，弱光照下的目标检测也是道路目标检测场景不得不面对的一个难题，本文没有针对弱光照场景进行优化，如何在黑暗环境下保障行车安全也将是一个重要的研究目标。

## 参考文献

[1] Wu B, Wan A, Iandola F, et al. SqueezeDet: Unified, Small, Low Power Fully Convolutional Neural Networks for Real-Time Object Detection for Autonomous Driving[J]. 2017 IEEE Conference on Computer Vision and Pattern Recognition Workshops (CVPRW), 2017.

[2] Karami E, Shehata M, Smith A. Image identification using SIFT algorithm: performance analysis against different image deformations[J]. arXiv preprint arXiv:1710.02728, 2017.

[3] Bay H, Tuytelaars T, Gool L V. Surf: Speeded up robust features: European conference on computer vision[C]: Springer, 2006.

[4] Dalal N, Triggs B. Histograms of oriented gradients for human detection: 2005 IEEE computer society conference on computer vision and pattern recognition (CVPR'05)[C]: Ieee, 2005.

[5] Krizhevsky A, Sutskever I, Hinton G E. Imagenet classification with deep convolutional neural networks[J]. Advances in neural information processing systems, 2012,25: 1097-1105.

[6] Girshick R, Donahue J, Darrell T, et al. Rich feature hierarchies for accurate object detection and semantic segmentation: Proceedings of the IEEE conference on computer vision and pattern recognition[C], 2014.

[7] Girshick R. Fast r-cnn: Proceedings of the IEEE international conference on computer vision[C], 2015.

[8] Ren S, He K, Girshick R, et al. Faster r-cnn: Towards real-time object detection with region proposal networks[J]. Advances in neural information processing systems, 2015,28: 91-99.

[9] Redmon J, Divvala S, Girshick R, et al. You only look once: Unified, real-time object detection: Proceedings of the IEEE conference on computer vision and pattern recognition[C], 2016.

[10] Redmon J, Farhadi A. YOLO9000: better, faster, stronger: Proceedings of the IEEE conference on computer vision and pattern recognition[C], 2017.

[11] Redmon J, Farhadi A. Yolov3: An incremental improvement[J]. arXiv preprint arXiv:1804.02767, 2018.

[12] Bochkovskiy A, Wang C, Liao H M. Yolov4: Optimal speed and accuracy of object detection[J]. arXiv preprint arXiv:2004.10934, 2020.

[13] Jocher G. yolov5. https://github.com/ultralytics/yolov5[Z]. 2021.

[14] Law H, Deng J. CornerNet: Detecting Objects as Paired Keypoints.[J]. International Journal of Computer Vision, 2020,3(128): 642-656.

[15] Duan K, Bai S, Xie L, et al. CenterNet: Keypoint Triplets for Object Detection: IEEE/CVF International Conference on Computer Vision (ICCV)[C], Seoul, Korea (South), 2019.

[16] Tian Z, Shen C, Chen H, et al. FCOS: Fully Convolutional One-Stage Object Detection, Seoul, Korea (South), 2019.

[17] Ge Z, Liu S, Li Z, et al. Ota: Optimal transport assignment for object detection: Proceedings of the IEEE/CVF Conference on Computer Vision and Pattern Recognition[C], 2021.

[18] Ge Z, Liu S, Wang F, et al. Yolox: Exceeding yolo series in 2021[J]. arXiv preprint arXiv:2107.08430, 2021.

[19] Yu J, Jiang Y, Wang Z, et al. UnitBox: An advanced object detection network: MM '16: Proceedings of the 24th ACM international conference on Multimedia[C], Amsterdam The Netherlands, 2016.

[20] Zheng Z, Wang P, Liu W, et al. Distance-IoU Loss: Faster and Better Learning for Bounding Box Regression: 34th AAAI Conference on Artificial Intelligence / 32nd Innovative Applications of Artificial Intelligence Conference / 10th AAAI Symposium on Educational Advances in Artificial Intelligence[C], New York, NY, 2020.

[21] Rezatofighi H, Tsoi N, Gwak J, et al. Generalized Intersection Over Union: A Metric and a Loss for Bounding Box Regression: 2019 IEEE/CVF Conference on Computer Vision and Pattern Recognition (CVPR)[C], Long Beach, CA, USA, 2019.

[22] Luo Z, Fang Z, Zheng S, et al. NMS-loss: learning with non-maximum suppression for crowded pedestrian detection: Proceedings of the 2021 International Conference on Multimedia Retrieval[C], 2021.

[23] Li J, Cheng B, Feris R, et al. Pseudo-IoU: Improving label assignment in anchor-free object detection: Proceedings of the IEEE/CVF conference on computer vision and pattern

recognition[C], 2021.